%% file: main.tex
\documentclass[10pt,twocolumn,letterpaper]{article}

\usepackage{cvpr}              

\input{preamble}

\definecolor{cvprblue}{rgb}{0.21,0.49,0.74}
\usepackage[pagebackref,breaklinks,colorlinks,citecolor=cvprblue]{hyperref}

\usepackage{graphicx}
\usepackage{subcaption}
\usepackage{tikz}
\usetikzlibrary{positioning}
\usepackage{url}

\def\paperID{*****} 
\def\confName{CVPR}
\def\confYear{2024}

\title{MagicVideo-V2: Multi-Stage High-Aesthetic Video Generation}

\author{
Weimin Wang\thanks{Equal contribution.} \and
Jiawei Liu\footnotemark[1] \and
Zhijie Lin \and
Jiangqiao Yan \and
Shuo Chen \and
Chetwin Low \and
Tuyen Hoang \and
Jie Wu \and
Jun Hao Liew \and
Hanshu Yan \and
Daquan Zhou \and
Jiashi Feng \and \\
Bytedance Inc.\\
{\tt\small \url{https://magicvideov2.github.io/}}
}

\begin{document}
\maketitle

\input{sec/0_abstract}    
\input{sec/1_intro}
\input{sec/2_method}
\input{sec/3_exp}

\input{sec/4_conclusion}
{
    \small
    \bibliographystyle{ieeenat_fullname}
    \bibliography{main}
}


\end{document}

%% file: preamble.tex
\usepackage[dvipsnames]{xcolor}


%% file: sec/0_abstract.tex
\begin{abstract}
The growing demand for high-fidelity video generation from textual descriptions has catalyzed significant research in this field. In this work, we introduce MagicVideo-V2 that  integrates  the  text-to-image model,  video motion generator, reference image embedding module and frame interpolation module into an end-to-end video generation pipeline. Benefiting from these architecture designs, MagicVideo-V2 can generate  an aesthetically pleasing, high-resolution video with remarkable fidelity and smoothness. It demonstrates superior performance over leading Text-to-Video systems such as Runway, Pika 1.0, Morph, Moon Valley and Stable Video Diffusion model via user evaluation at large scale. 
\end{abstract}

%% file: sec/1_intro.tex
\section{Introduction}
\label{sec:intro}
The proliferation of Text-to-Video (T2V) models has marked a significant advancement~\cite{zhou2023magicvideo, girdhar2023emu, blattmann2023stable, kondratyuk2023videopoet}, driven by the recent diffusion-based  models. In this work, we propose MagicVideo-V2, a novel multi-stage T2V framework that integrates Text-to-Image (T2I), Image-to-Video (I2V), Video-to-Video (V2V) and Video Frame Interpolation (VFI) modules into an end-to-end video generation pipeline.


The T2I module sets the foundation by producing an initial image from the text prompt, capturing the aesthetic essence of the input. Then the I2V module takes the image as input and outputs  low-resolution keyframes of the generated video. The subsequent V2V module  increases the resolution of the keyframes and enhances their details. Finally, the frame interpolation module    adds smoothness to the motion in the video.

\begin{figure*}[t!]
\centering
    \includegraphics[width=.95\linewidth]{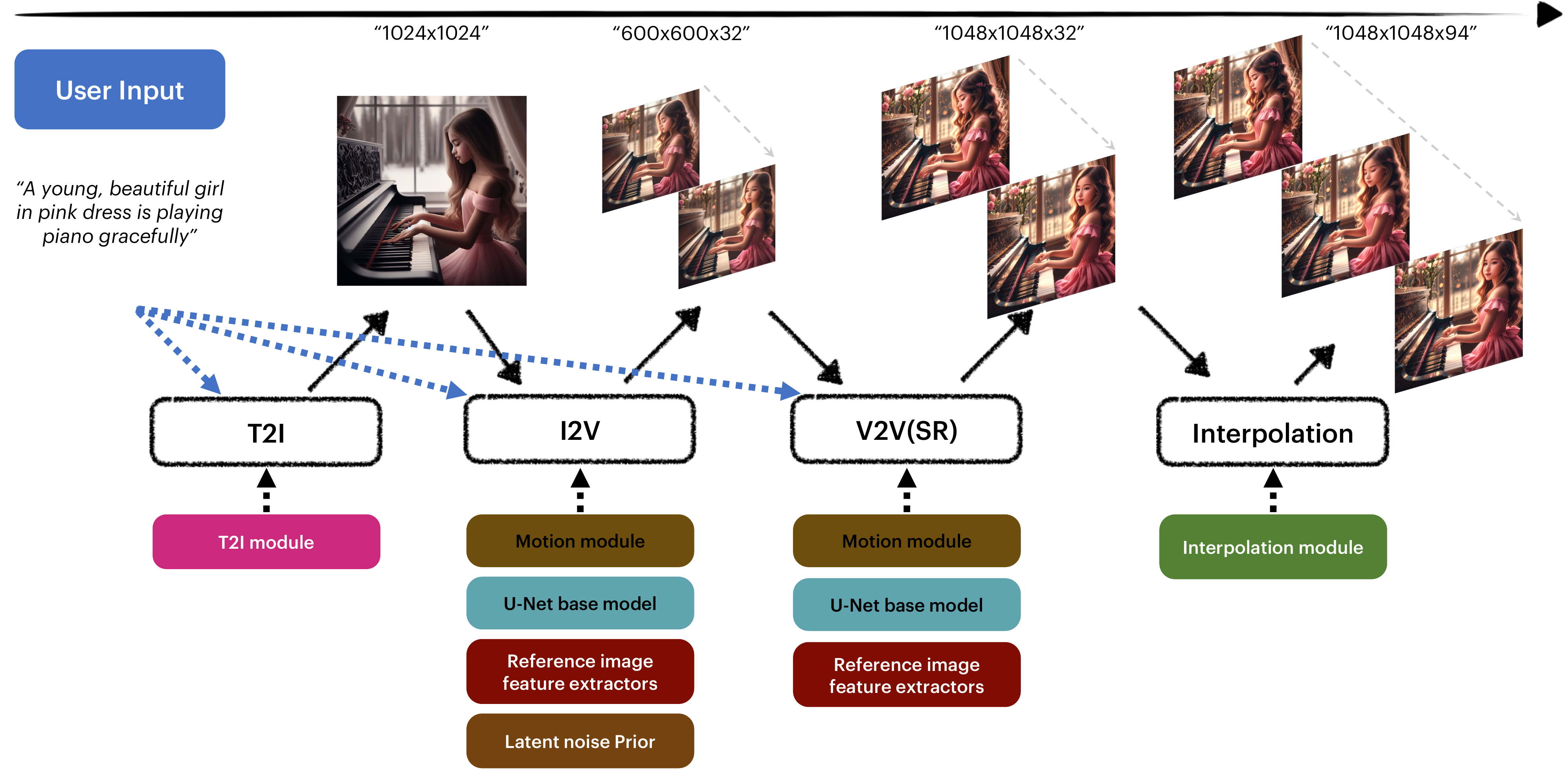}
    \caption{Overview of MagicVideo-V2. The T2I module creates a 1024$\times$1024 image that encapsulates the described scene.  Subsequently, the I2V module animates this still image, generating a sequence of 600$\times$600$\times$32 frames, with the latent noise prior ensuring continuity from the initial frame. The V2V module   enhances these frames to a 1048$\times$1048 resolution while refining the video content. Finally, the interpolation module extends the sequence to 94 frames, getting a 1048$\times$1048 resolution video that exhibits both high aesthetic quality and temporal smoothness.}
\end{figure*}

%% file: sec/2_method.tex
\section{MagicVideo-V2}
\label{sec:method}
The proposed MagicVideo-V2 is a multi-stage   end-to-end video generation pipeline capable of generating high-aesthetic videos from textual description. It consists of the following key modules:
\begin{itemize}
    \item \textbf{Text-to-Image} model that generates an aesthetic image with high fidelity from the given text prompt.
    \item \textbf{Image-to-Video} model that uses the text prompt and generated image as conditions to generate keyframes.
    \item \textbf{Video to video} model that refines and performs super-resolution on the keyframes to yield a high-resolution video.
    \item \textbf{Video Frame Interpolation} model that interpolates frames between  keyframes to smoothen the video motion and  finally generates a high resolution, smooth, highly aesthetic video.
\end{itemize}
The following subsections will explain each module  in more details.

\subsection{The Text-to-Image Module}
\label{subsec:T2I}
The T2I module takes a text prompt from users as input and generates a 1024 $\times$ 1024 image as  the reference image for video generation. The reference image helps describe the video contents and the aesthetic style. 
The proposed MagicVideo-V2 is compatible with  different T2I models. 
Specifically, we use a internally developed  diffusion-based T2I model in MagicVideo-V2 that could output high aesthetic images.

\subsection{The Image-to-Video Module}
\label{subsec:I2V}
The I2V module is built on a high-aesthetic SD1.5 \cite{rombach2022highresolution}  model, that leverages human feedback to   improve model capabilities in   visual quality an content consistency.
The I2V module inflates this  high-aesthetic SD1.5  with a motion module inspired by \cite{guo2023animatediff}, both of which were trained on   internal datasets. 

The I2V module is augmented with a reference image embedding module for utilizing the reference image.
More specifically, we adapt an appearance encoder to extract the reference image embeddings and inject them into the I2V module via a cross-attention mechanism. 
In this way, the image prompt can be effectively decoupled   from the text prompts and provide stronger image  conditioning. 
In addition, we employ a latent noise prior strategy to provide layout condition in the starting noisy latents. The frames are initialized from standard Gaussian noise whose means have shifted from zeros towards the value of reference image latent. With a proper noise prior trick, the image layout could be partially retained and the temporal coherence across frames could also be improved. 
To further enhance layout and spatial conditioning, we deploy a ControlNet \cite{zhang2023adding} module to directly extract RGB information from the reference image and apply it to all frames. 
These techniques align the  the frames with the reference image  well while allowing the model to generate clear motion. 


We employ an image-video joint training strategy for training the I2V module, 
where the images are treated as single-frame videos. The motivation here for joint training is to leverage our internal image datasets of high quality and aesthetics, to improve frame quality of generated videos. The image dataset part also serves as a good compensation for our video datasets that are lacking in diversity and volume.

\subsection{The Video-to-Video Module}
\label{subsec:VSR}
The V2V module has a similar design as the  I2V module. It shares the same backbone and spatial layers as in I2V module. Its motion module   is separately finetuned using a high-resolution video  subset  for video super-resolution. 

The  image apperance encoder  and ControlNet module are also used here. This turns out to be crutial, as we are generating video frames at a much higher resolution. Leveraging the information from the   reference image helps guide the video diffusion steps by reducing structural errors and failure rates. In addition, it could also enhance the details generated at the higher resolution. 


\subsection{Video Frame Interpolation (VFI)}

The VFI module uses an internally trained GAN based VFI model. It employs an Enhanced Deformable Separable Convolution (EDSC) head \cite{cheng2021multiple} paired with a VQ-GAN based architecture, similar to the autoencoder model used in the research conducted by \cite{danier2023ldmvfi}. To further enhance its stability and smoothness, we used a pretrained lightweight interpolation model proposed in \cite{zhang2023extracting}. 

%% file: sec/3_exp.tex
\section{Experiment}
\label{sec:exp}

\subsection{Human evaluations}
\begin{figure}
\centering
    \includegraphics[width=\linewidth]{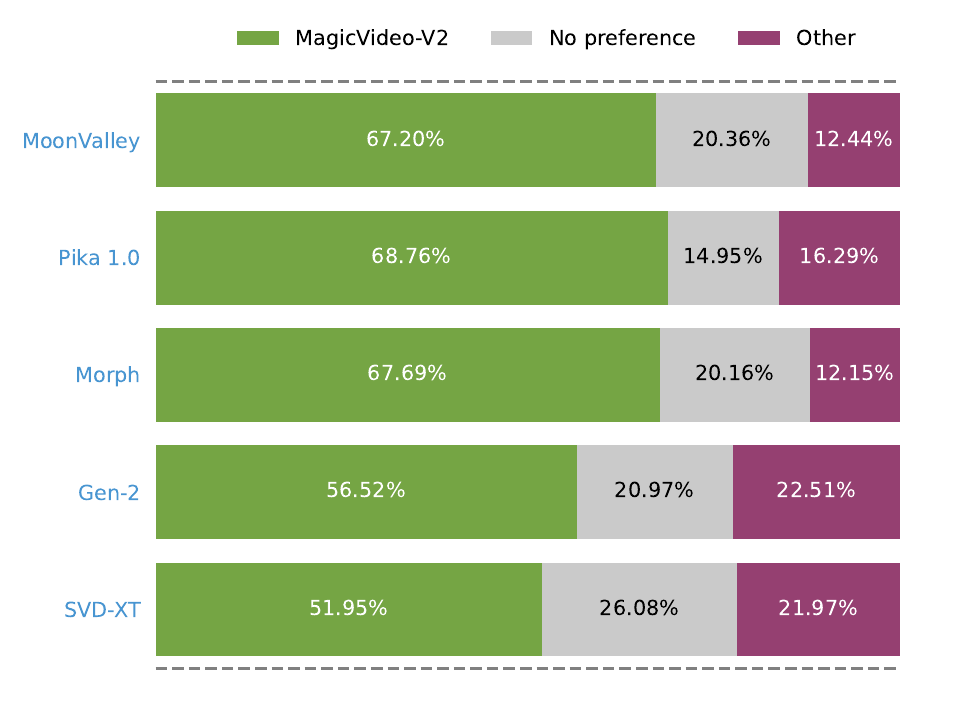}
    \caption{
    The distribution of human evaluators' perferences, showing a dominant inclination towards MagicVideo-V2 over other state-of-the-art T2V methods.
    Green, gray, and pink bars represent trials where MagicVideo-V2 was judged better, equivalent, or inferior, respectively.}
    \label{fig:preference}
\end{figure}

To evaluate MagicVideo-V2, we engaged human evaluators to conduct comparative analyses with contemporary state-of-the-art T2V systems. A panel of 61 evaluators rated 500 side-by-side comparisons between MagicVideo-V2 and an alternative T2V method. Each voter is presented with a random pair of videos, including one of ours vs one of the competitors, based on the same text prompt, for each round of comparison. They were presented with three assessment options-Good, Same, or Bad-indicating a preference for MagicVideo-V2, no preference, or a preference for the competing T2V method, respectively. The voters are requested to cast their vote based on their overall preference on three criteria: 1) which video has higher frame quality and overall visual appealing. 2) which video is more temporal consistent, with better motion range and motion validility. 3) which video has fewer structure errors, or bad cases. 
The compiled statistics of these trials can be found in Table~\ref{tab:gsb}, with the proportions of preferences depicted in Figure~\ref{fig:preference}.
The results demonstrate a clear preference for MagicVideo-V2, evidencing its superior performance from the standpoint of human visual perception.

\begin{table}[htb]
    \resizebox{\linewidth}{!}{
    \centering
    \begin{tabular}{@{}lcccc@{}}
        \toprule
         Method & Good (G) & Same (S) & Bad (B) & (G+S)/(B+S) \\
         \midrule
         MoonValley~\cite{moonvalley} & 4099 & 1242 & 759 & 2.67 \\
         Pika 1.0~\cite{pika} & 4263 & 927 & 1010 & 2.68 \\
         Morph~\cite{morph} & 4129 & 1230 & 741 & 2.72 \\
         Gen-2~\cite{gen2} & 3448 & 1279 & 1373 & 1.78 \\
         SVD-XT~\cite{svd-xt} & 3169 & 1591 & 1340 & 1.62 \\
         \bottomrule
    \end{tabular}
    }
    \caption{Human side-by-side evaluations comparing MagicVideo-V2 with other state-of-the-art text-to-video generation methods, indicating a strong preference for MagicVideo-V2. }
    \label{tab:gsb}
\end{table}

\subsection{Qualitative examples}

\begin{figure}[htb!]
  \begin{subfigure}[b]{\linewidth}
    \centering
    \includegraphics[width=0.24\textwidth]{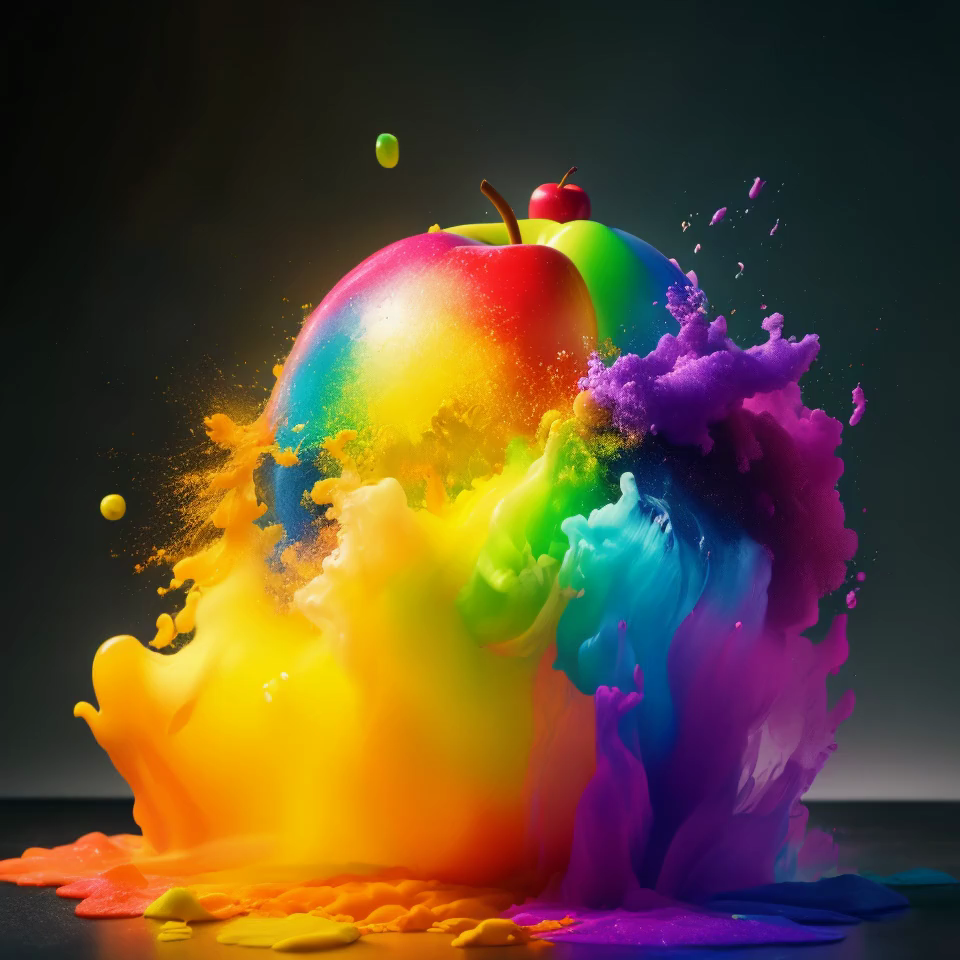}\hfill
    \includegraphics[width=0.24\textwidth]{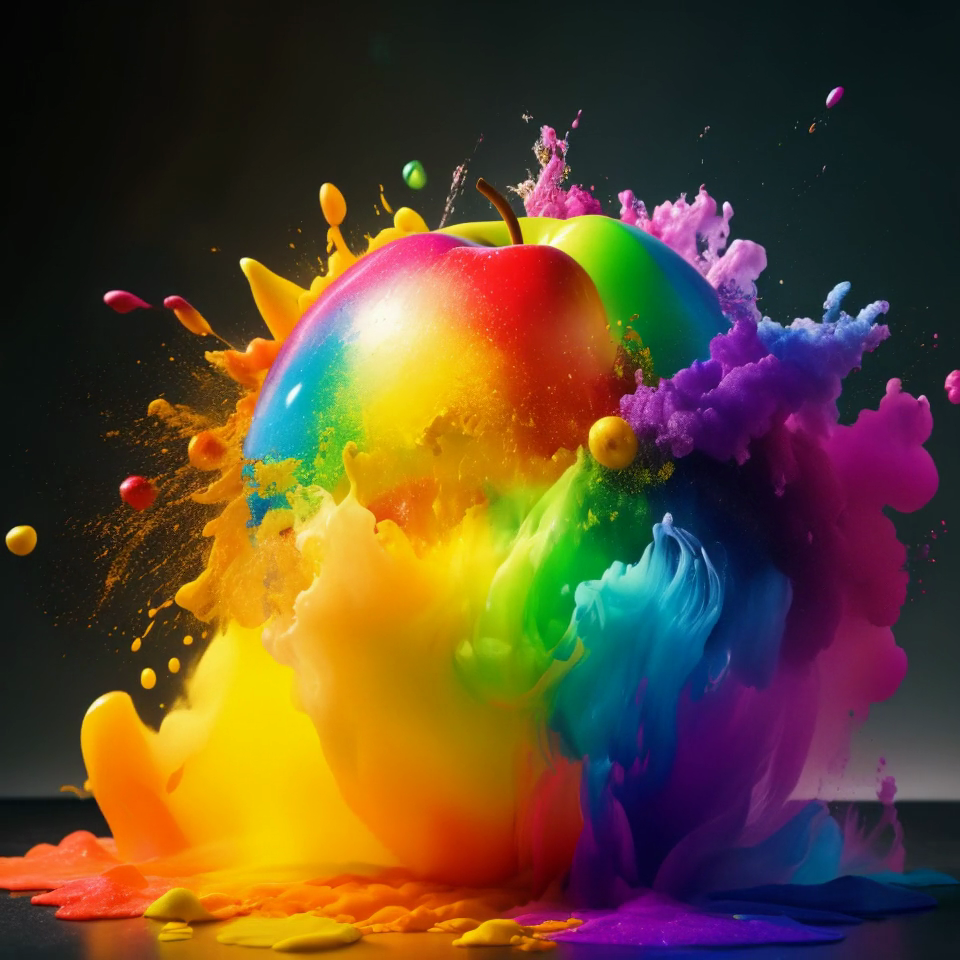}\hfill
    \includegraphics[width=0.24\textwidth]{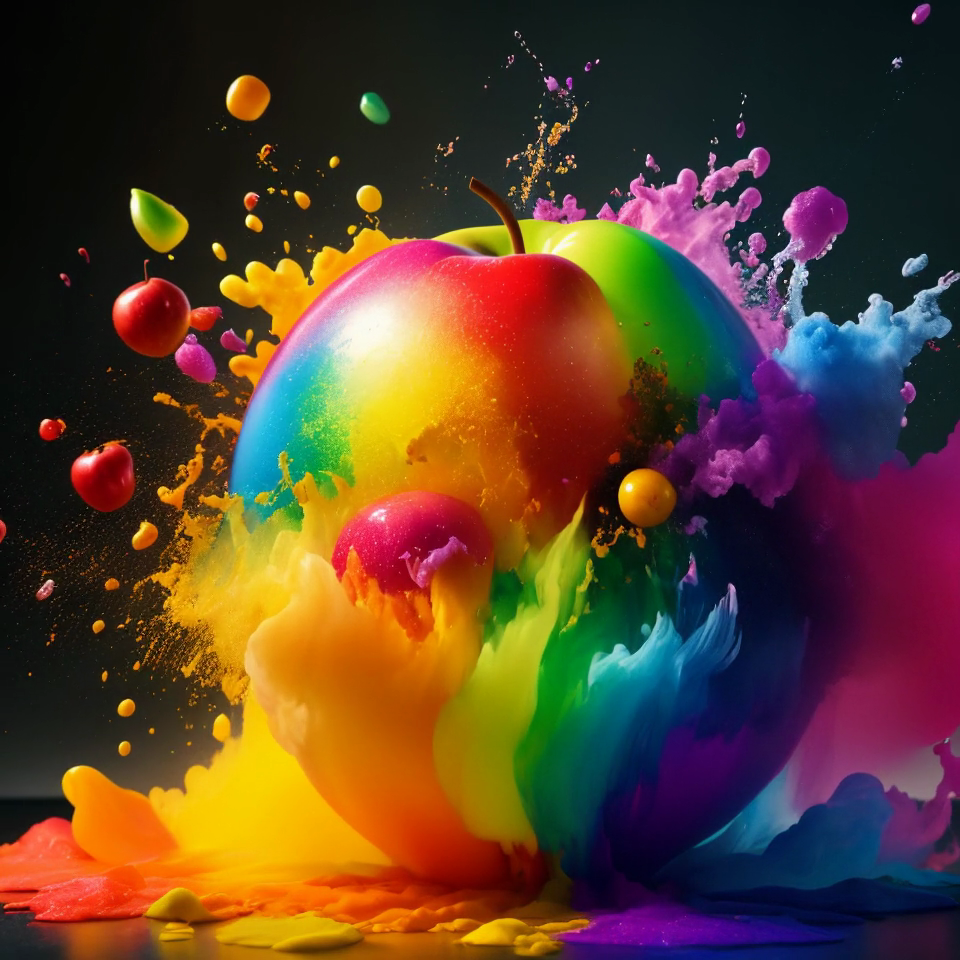}\hfill
    \includegraphics[width=0.24\textwidth]{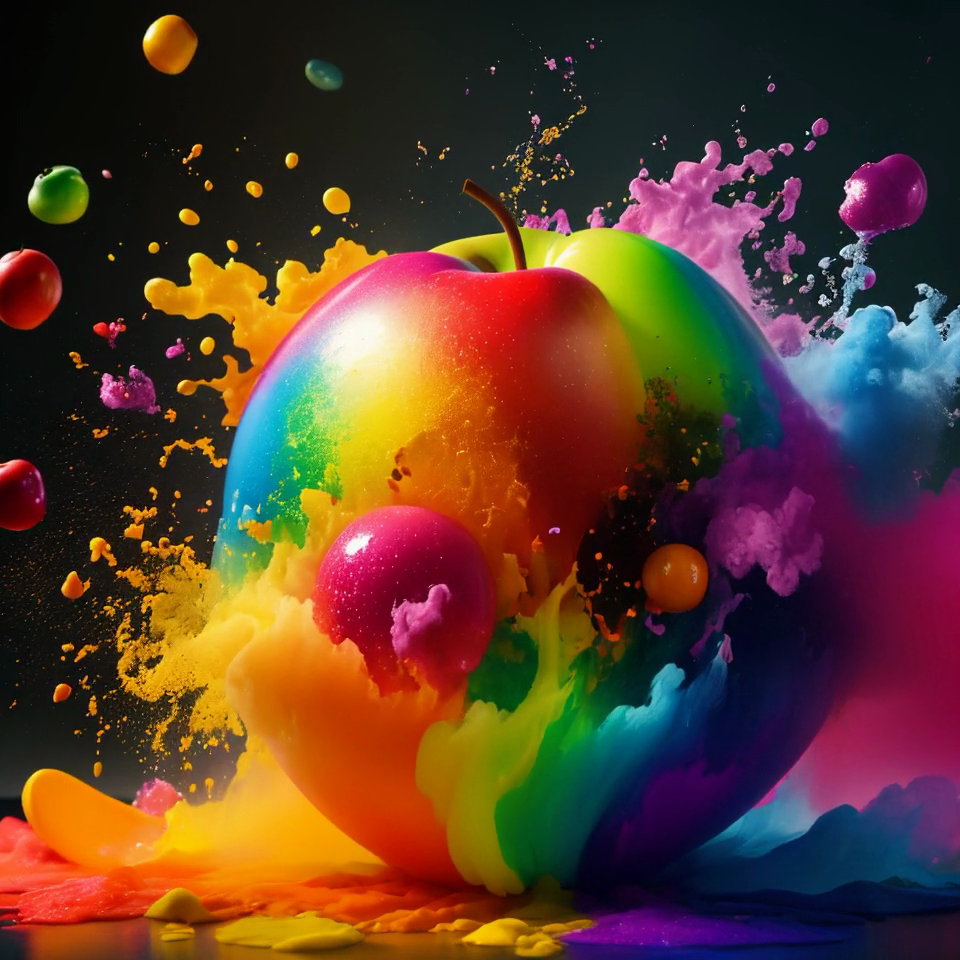}
    \caption*{Prompt: A large blob of exploding splashing rainbow paint, with an apple emerging, 8k.}
  \end{subfigure}
  
  \vspace{1em}

  \begin{subfigure}[b]{\linewidth}
    \centering
    \includegraphics[width=0.24\textwidth]{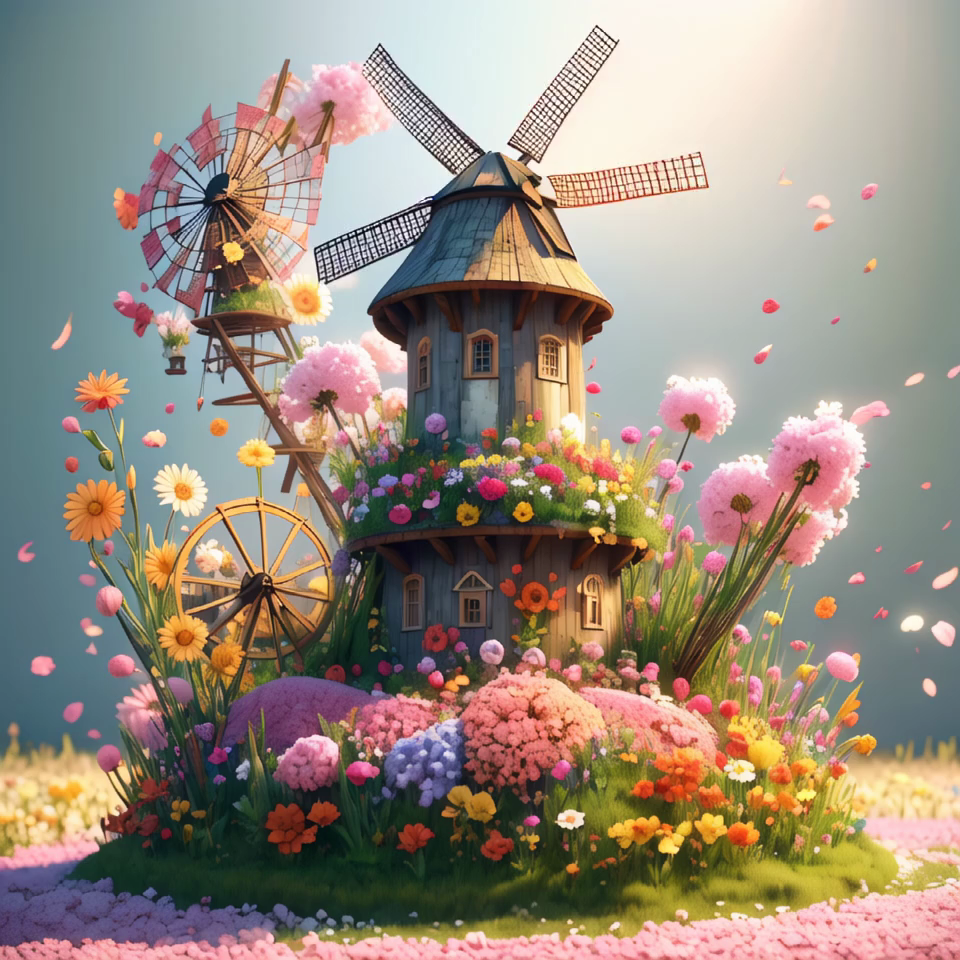}\hfill
    \includegraphics[width=0.24\textwidth]{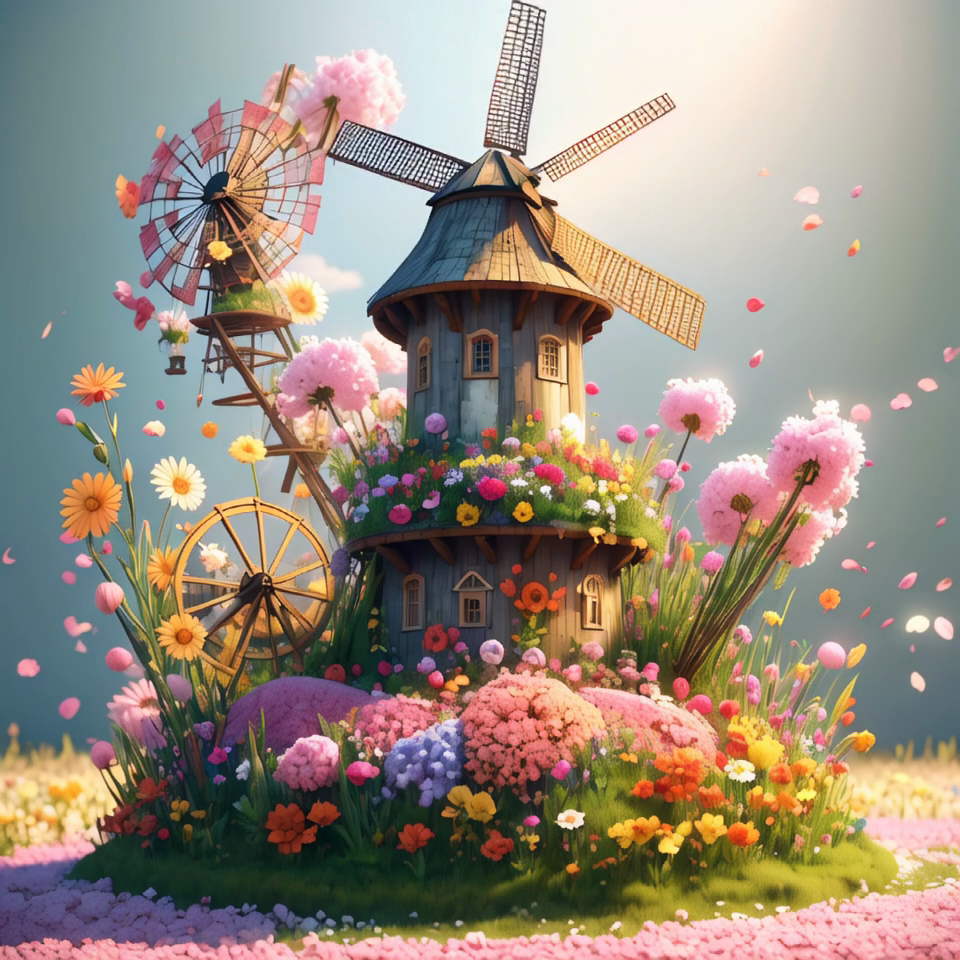}\hfill
    \includegraphics[width=0.24\textwidth]{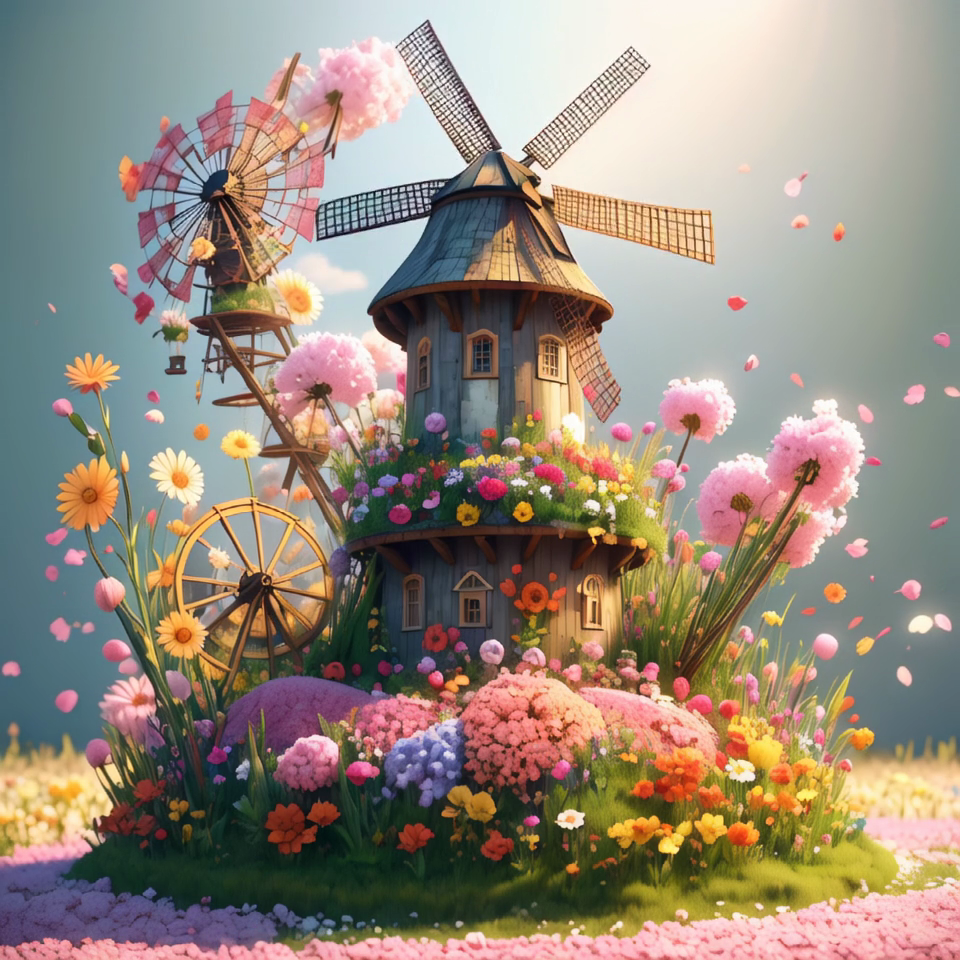}\hfill
    \includegraphics[width=0.24\textwidth]{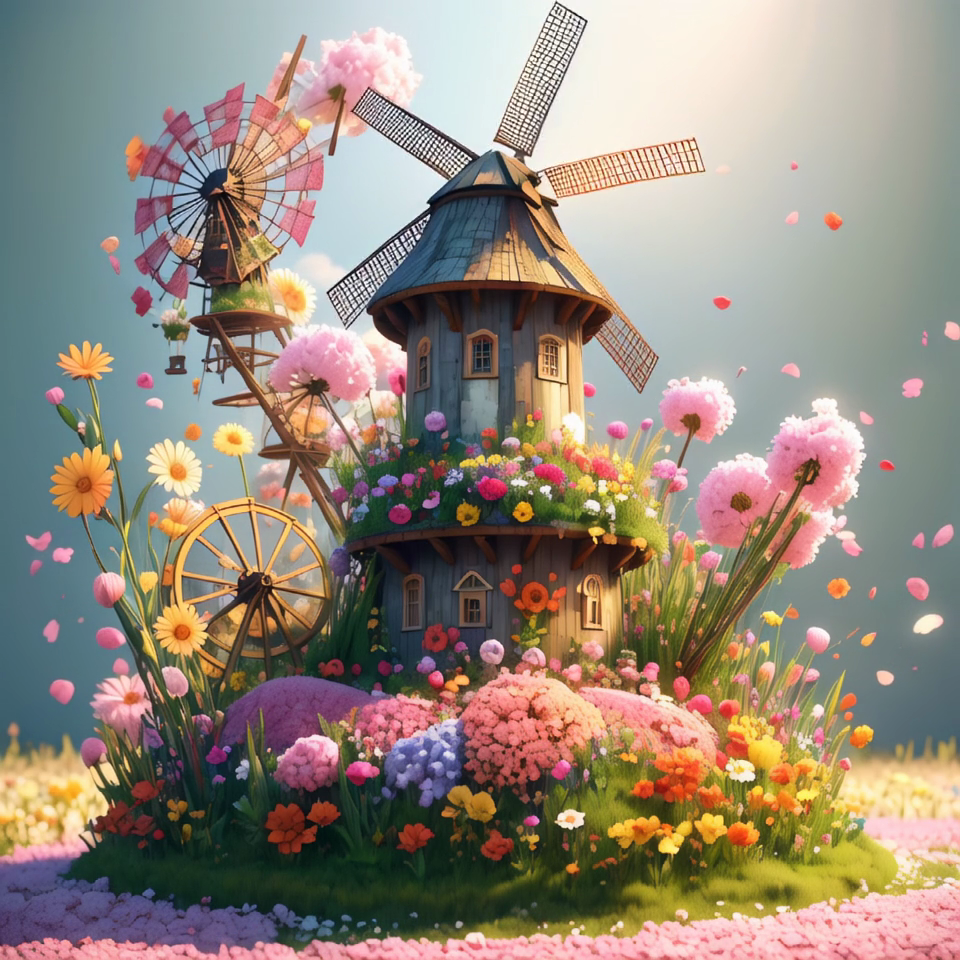}
    \caption*{Prompt: An old-fashioned windmill surrounded by flowers, 3D design.}
  \end{subfigure}
  
  \vspace{1em}

  \begin{subfigure}[b]{\linewidth}
    \centering
    \includegraphics[width=0.24\textwidth]{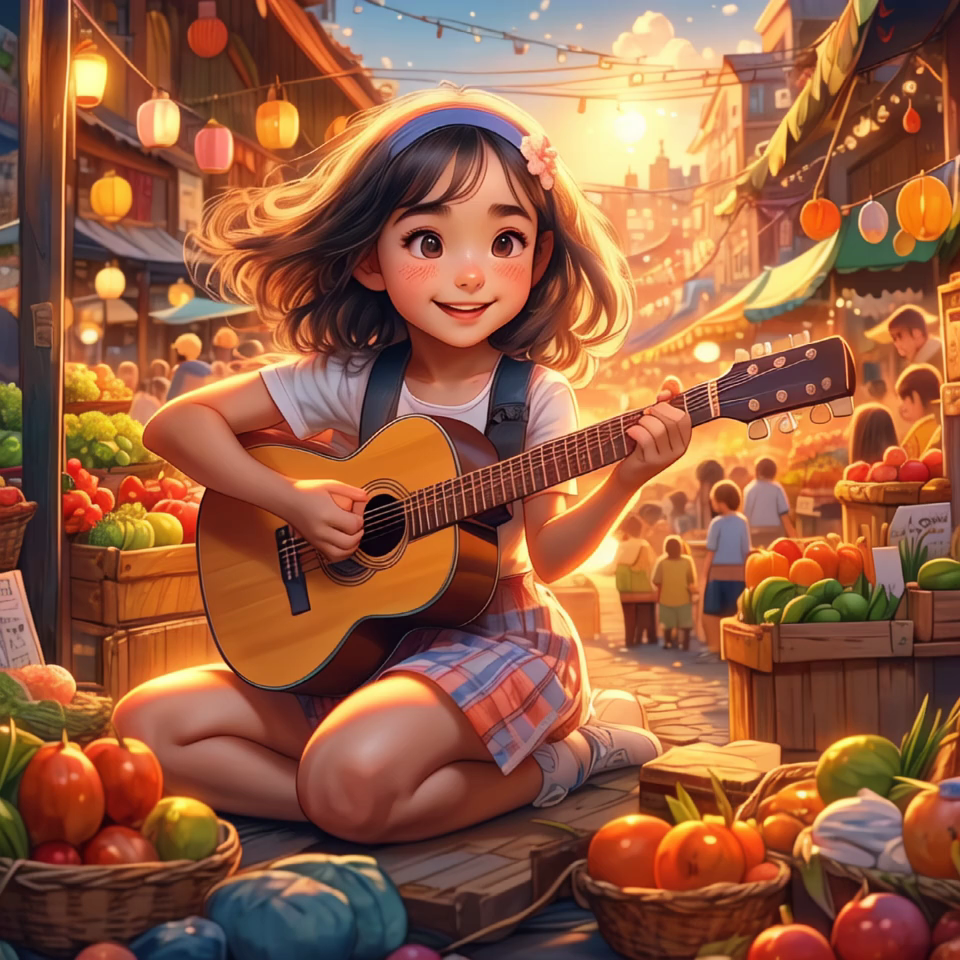}\hfill
    \includegraphics[width=0.24\textwidth]{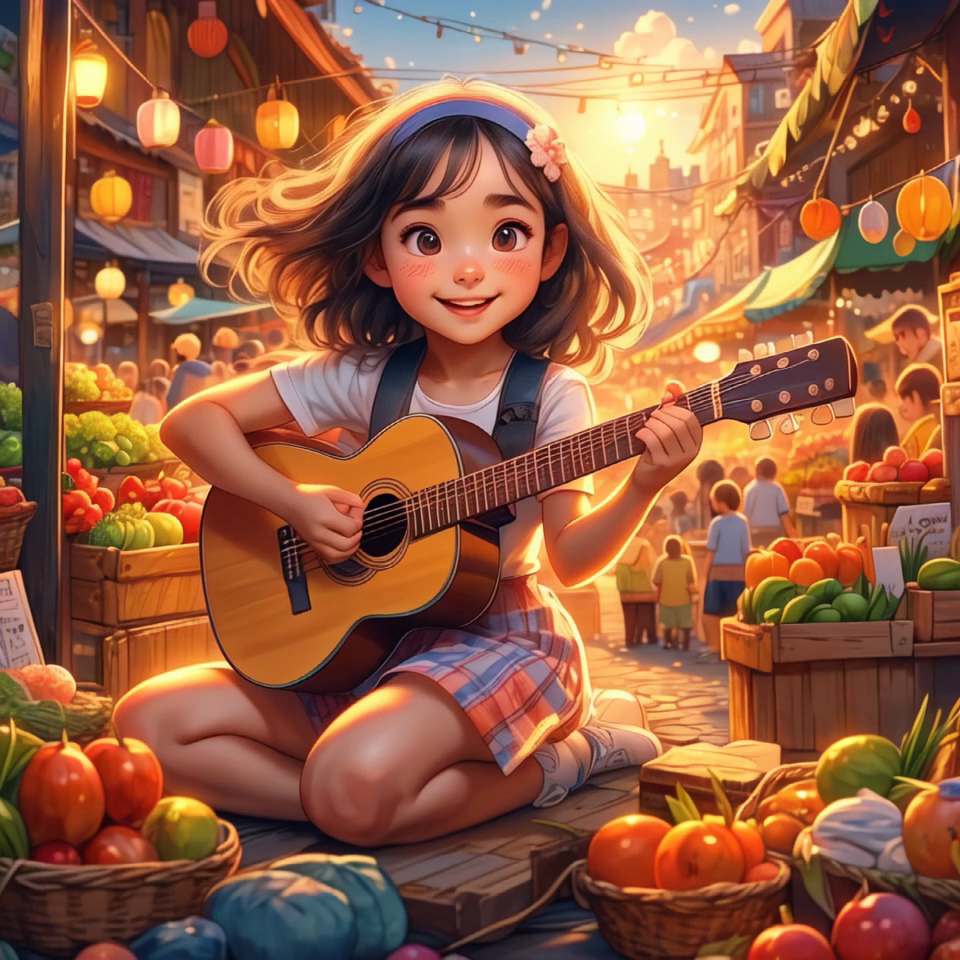}\hfill
    \includegraphics[width=0.24\textwidth]{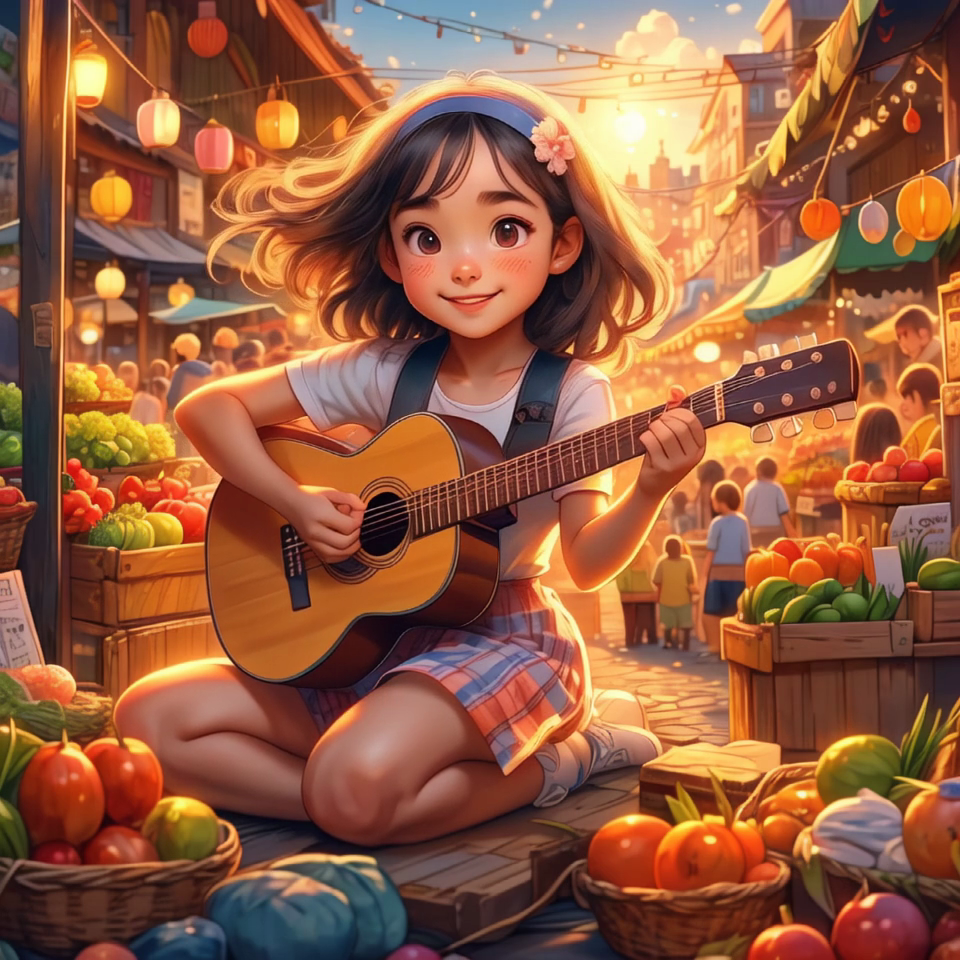}\hfill
    \includegraphics[width=0.24\textwidth]{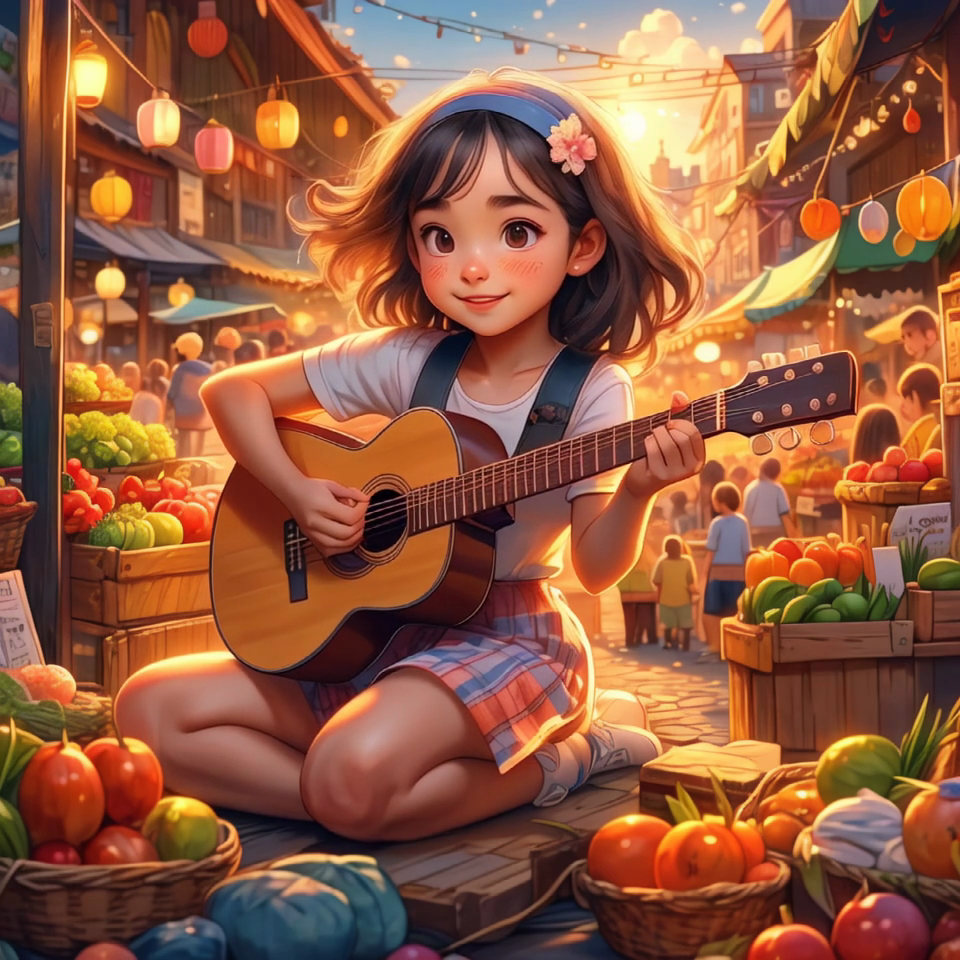}
    \caption*{Prompt: A girl with a hairband performing a song with her guitar on a warm evening at a local market, children's story book.}
  \end{subfigure}
  
  \vspace{1em}
  
  \begin{subfigure}[b]{\linewidth}
    \centering
    \includegraphics[width=0.24\textwidth]{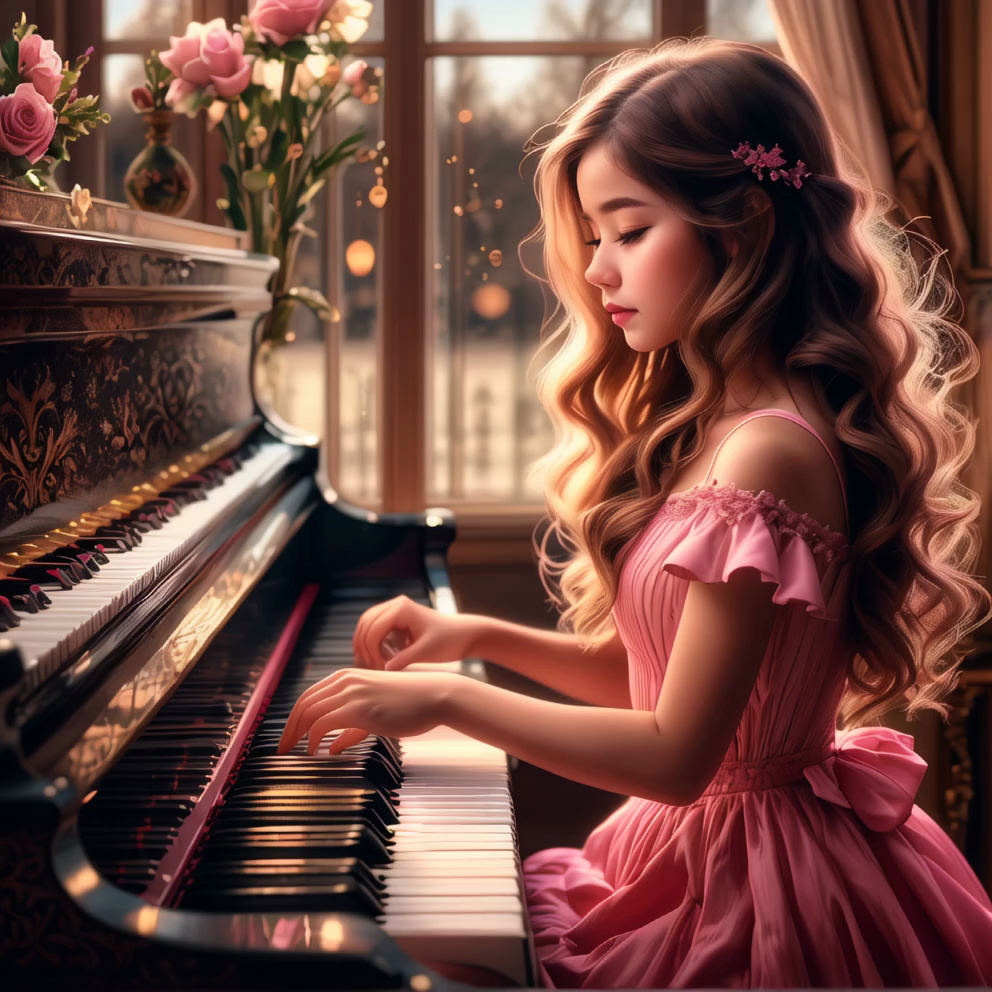}\hfill
    \includegraphics[width=0.24\textwidth]{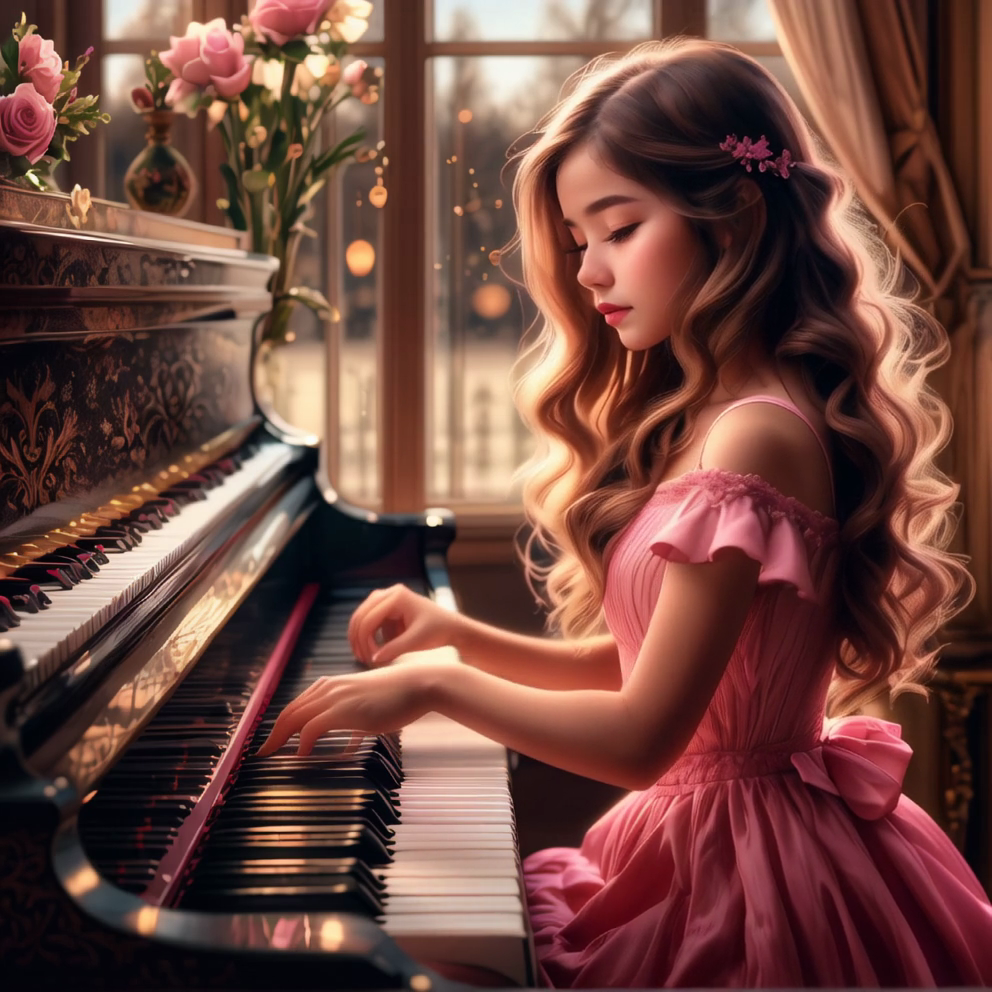}\hfill
    \includegraphics[width=0.24\textwidth]{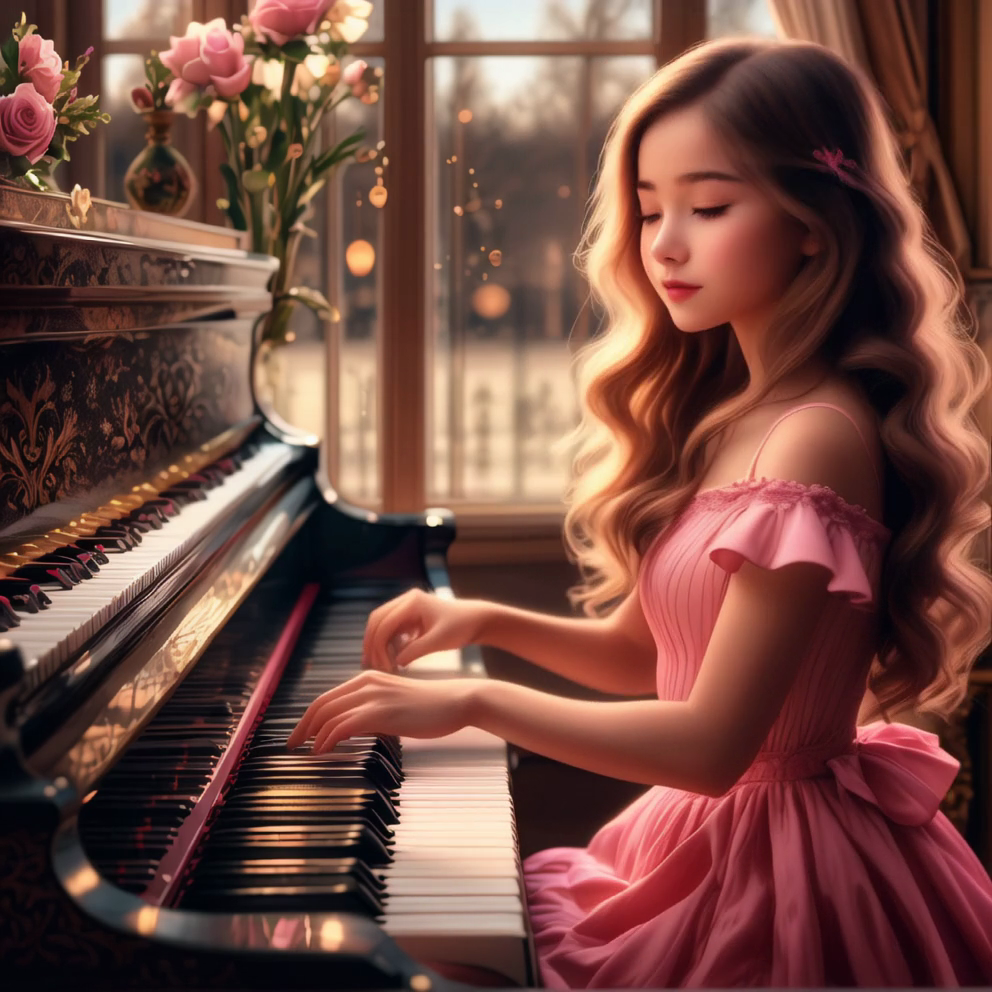}\hfill
    \includegraphics[width=0.24\textwidth]{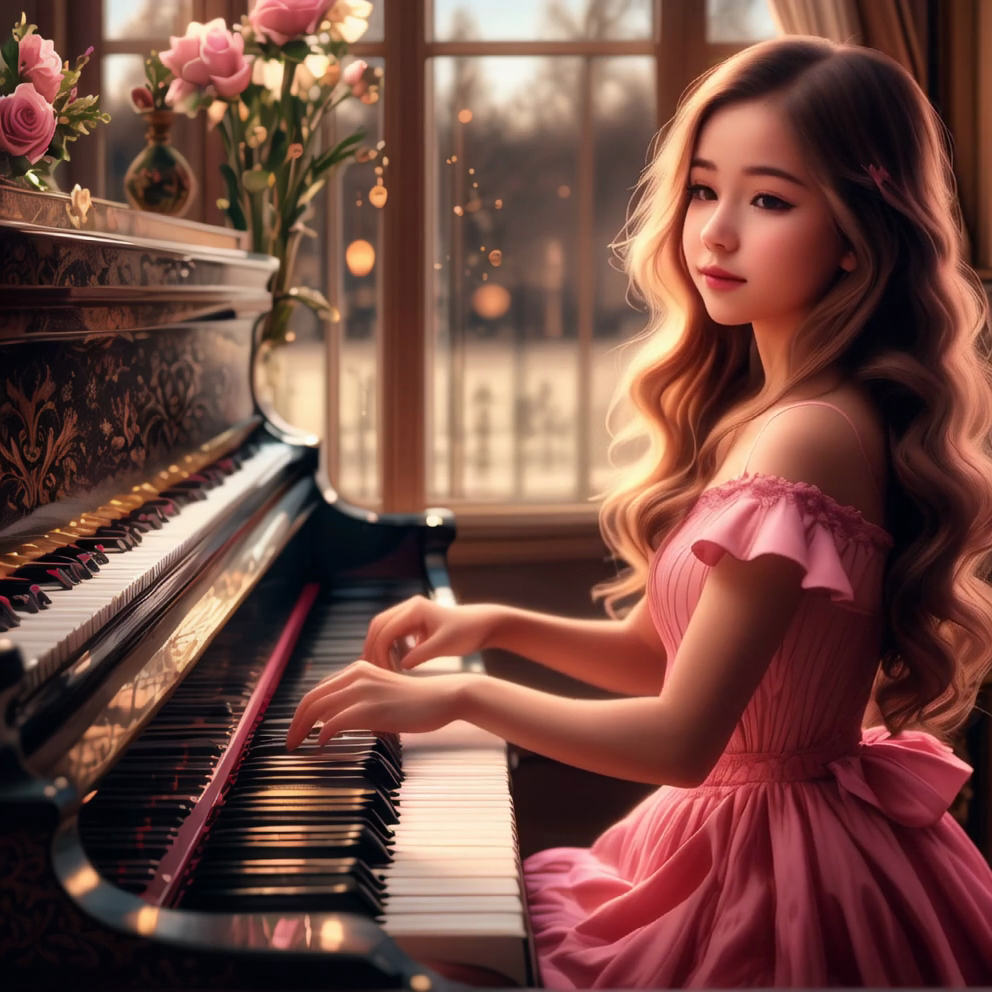}
    \caption*{Prompt: A young, beautiful girl in a pink dress is playing piano gracefully.}
  \end{subfigure}
  
\caption{Examples of MagicVideo-V2 generated videos via a text prompt.}
\label{fig:example}
\end{figure}

\begin{figure*}[htb!]
\centering
  \begin{subfigure}[b]{\linewidth}
    \centering
    \raisebox{-.5\height}{\includegraphics[width=0.15\textwidth]{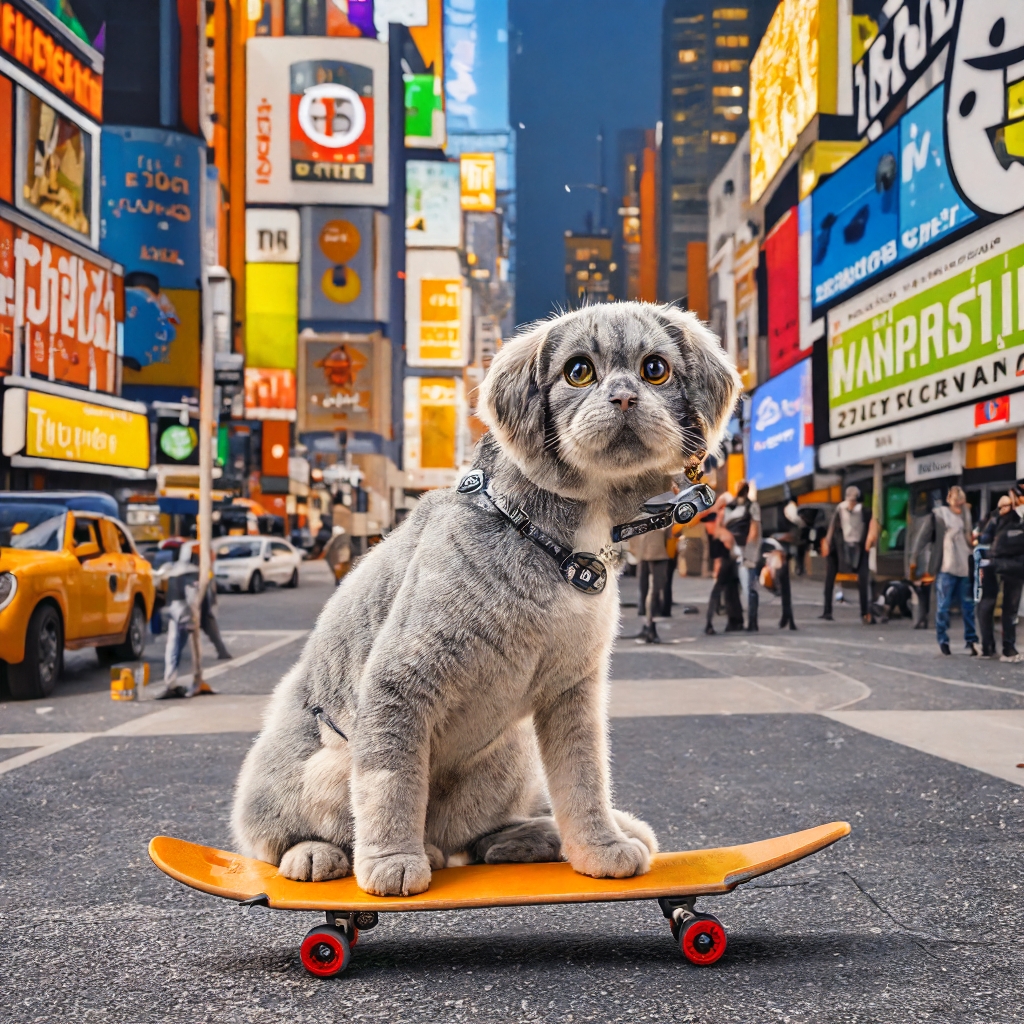}}
    \raisebox{-.5\height}{
    \begin{tikzpicture}
    \draw[->, line width=2pt] (0,0) -- (0.5,0);
    \end{tikzpicture}
    }
    \raisebox{-.5\height}{\includegraphics[width=0.25\textwidth]{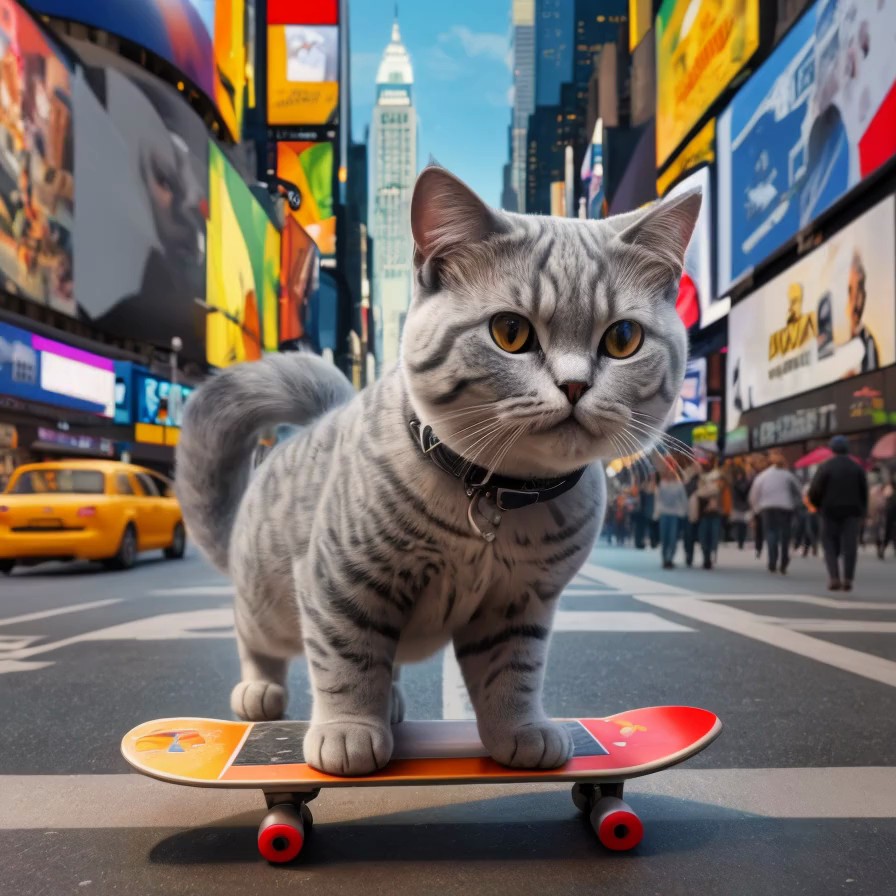}}
    \raisebox{-.5\height}{\includegraphics[width=0.25\textwidth]{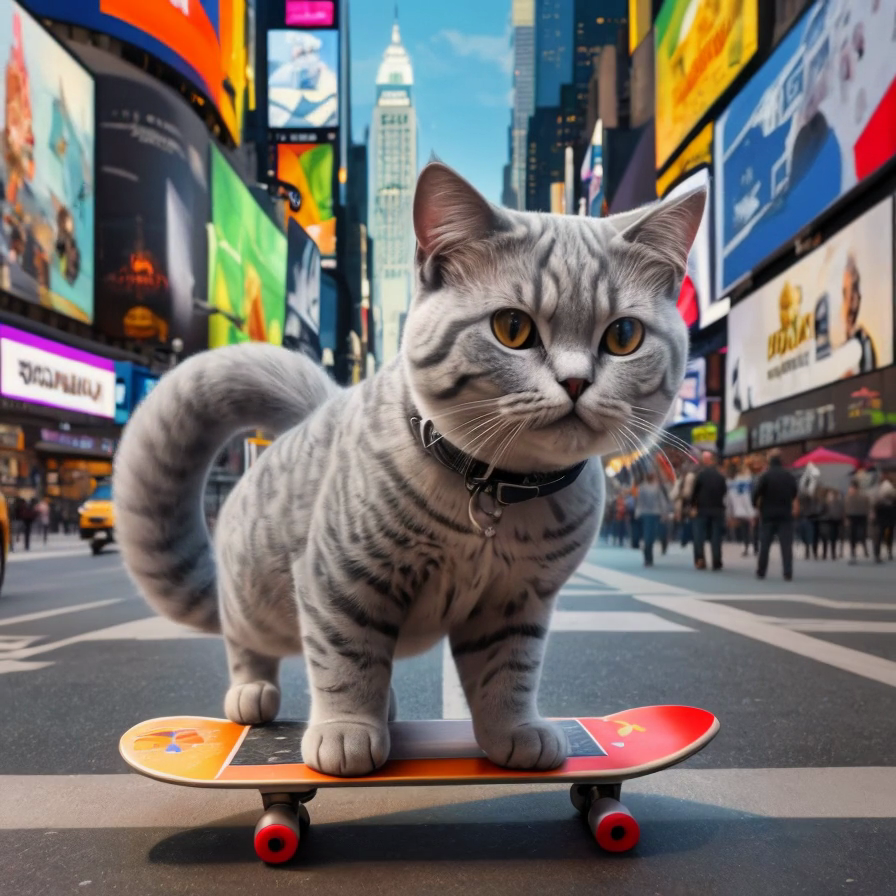}}
    \raisebox{-.5\height}{\includegraphics[width=0.25\textwidth]{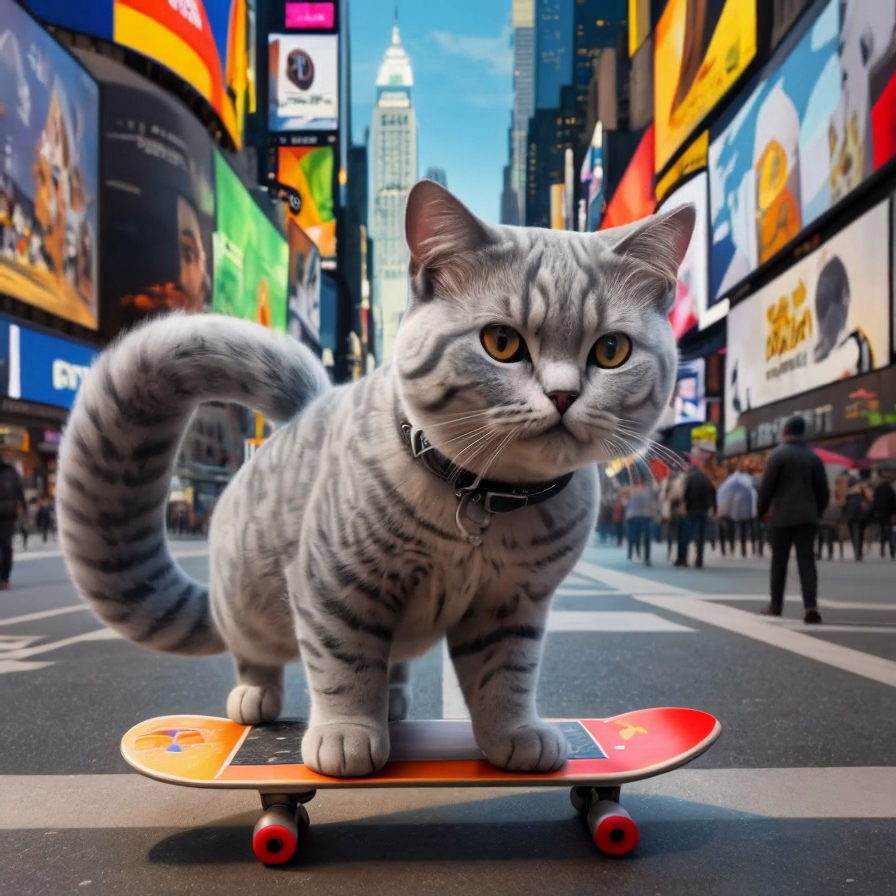}}
    \caption*{Prompt: ``A gray British \textbf{Shorthair} skateboarding in Times Square, in cubist painting style.'' The wrong dog generated from the T2I module is fixed by the I2V and V2V module.}
  \end{subfigure}
  \vspace{.5em}
  
  \begin{subfigure}[b]{\linewidth}
    \centering
    \raisebox{-.5\height}{\includegraphics[width=0.15\textwidth]{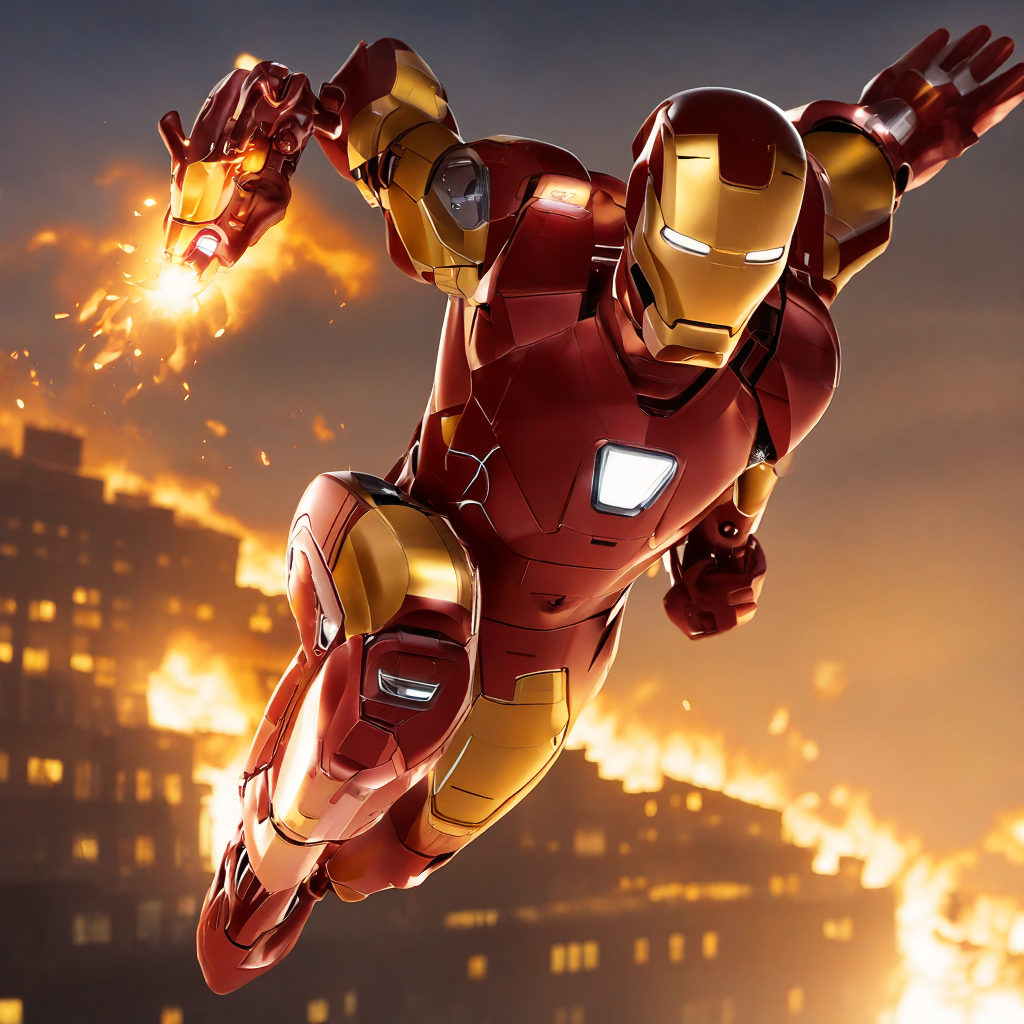}}
    \raisebox{-.5\height}{
    \begin{tikzpicture}
    \draw[->, line width=2pt] (0,0) -- (0.5,0);
    \end{tikzpicture}
    }
    \raisebox{-.5\height}{\includegraphics[width=0.25\textwidth]{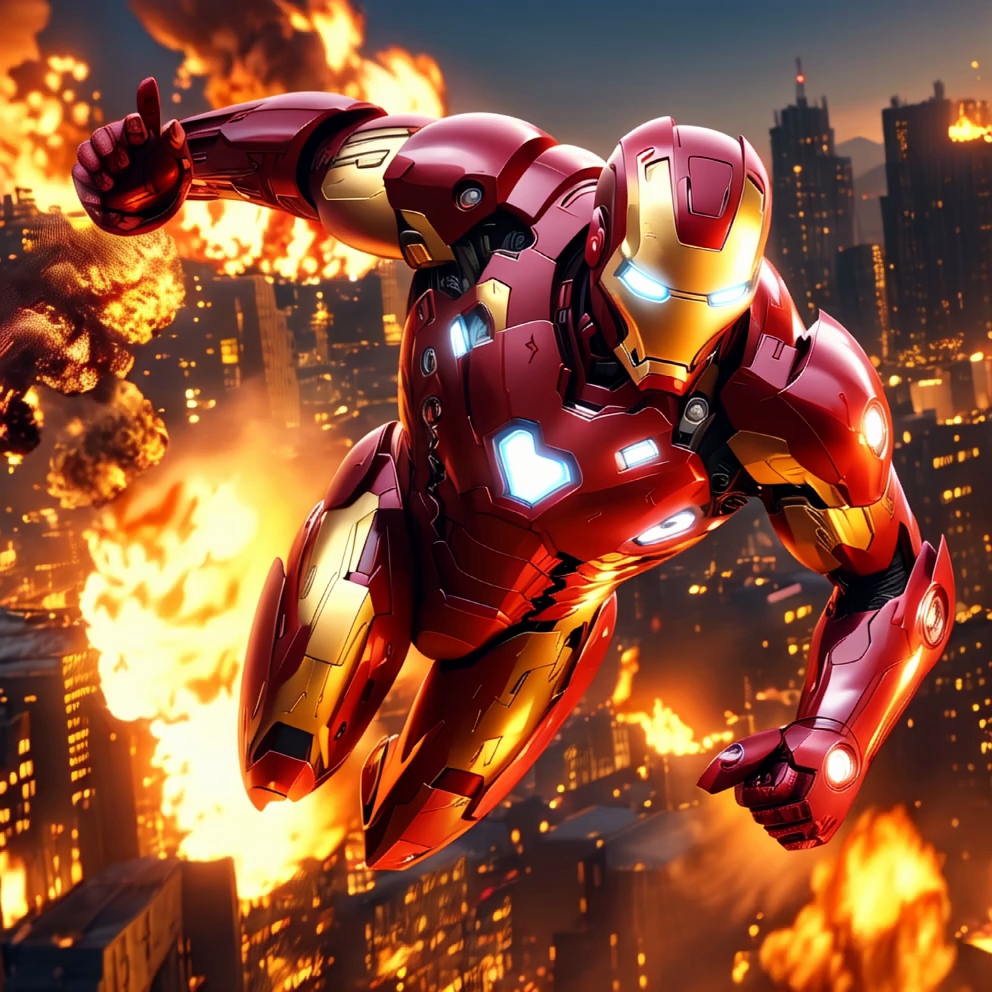}}
    \raisebox{-.5\height}{\includegraphics[width=0.25\textwidth]{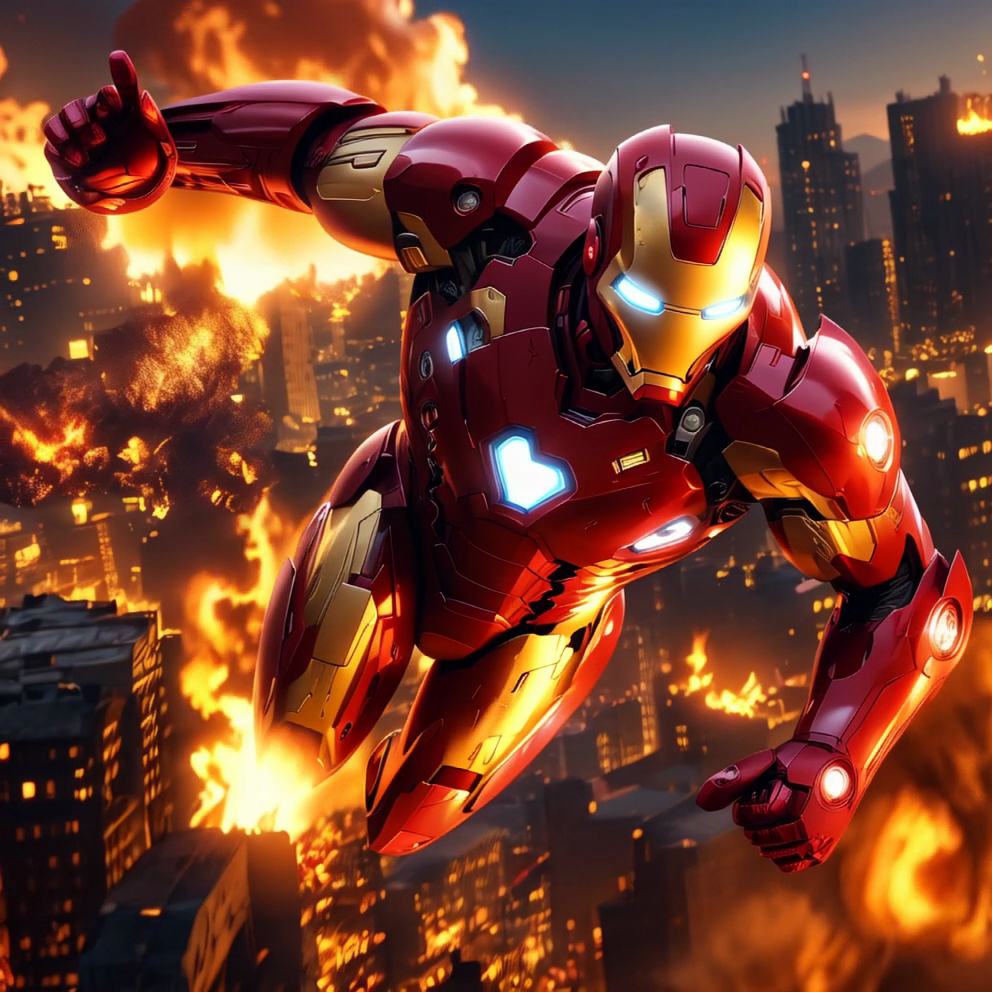}}
    \raisebox{-.5\height}{\includegraphics[width=0.25\textwidth]{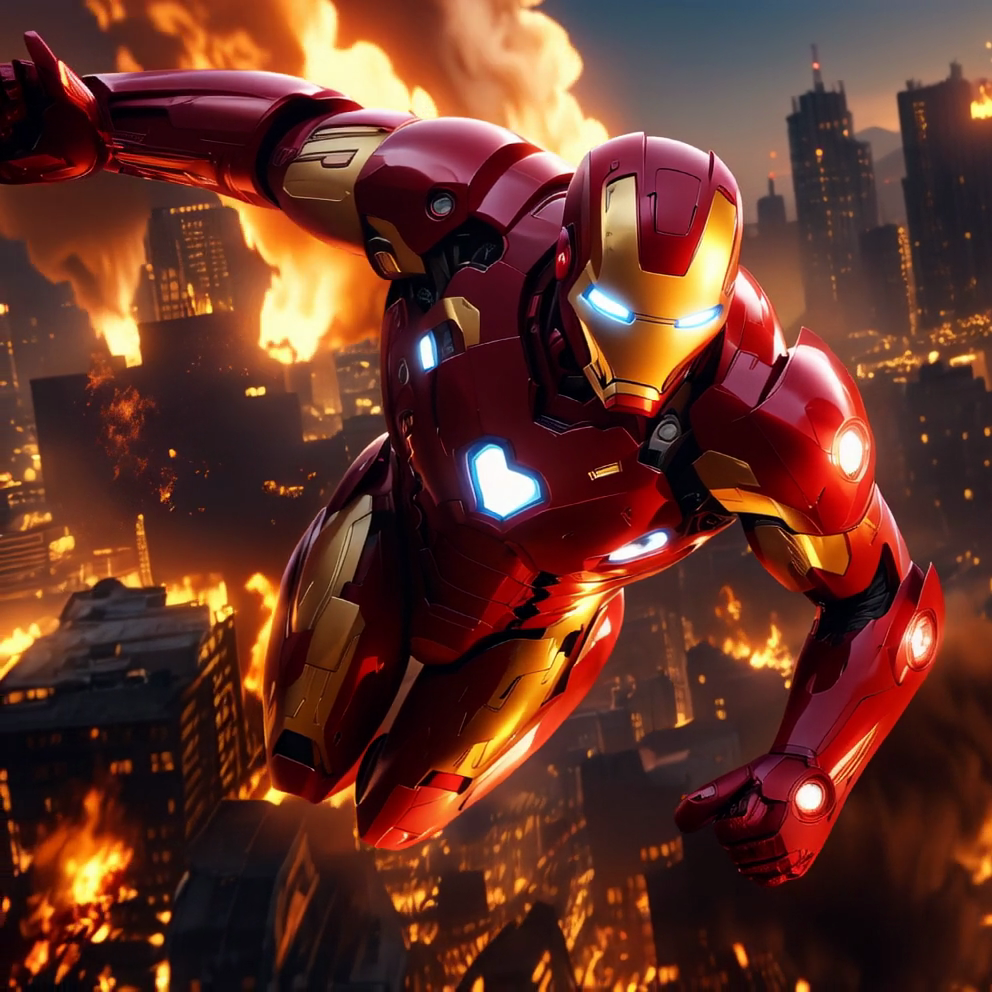}}
    \caption*{Prompt: ``Ironman flying over a burning city, very detailed surroundings, cities are blazing, shiny iron man suit, realistic, 4k ultra high defi'' The ironman's redundant arm is removed by the I2V and V2V module.}
  \end{subfigure}
  \vspace{.5em}
  
  \begin{subfigure}[b]{\linewidth}
    \centering
    \raisebox{-.5\height}{\includegraphics[width=0.15\textwidth]{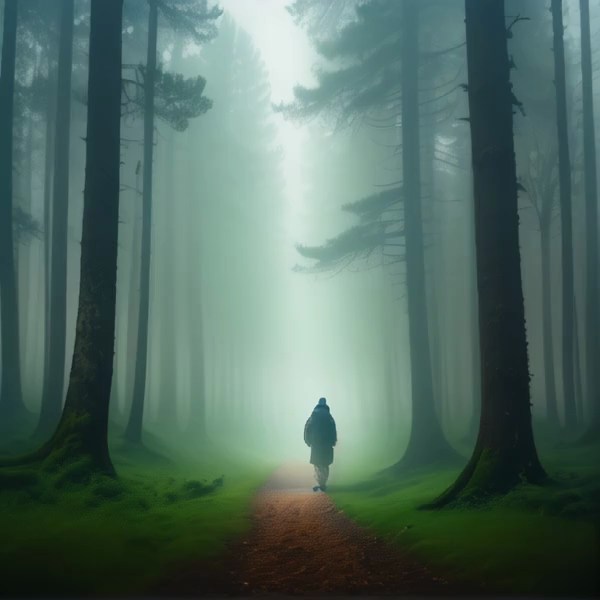}}
    \raisebox{-.5\height}{\includegraphics[width=0.15\textwidth]{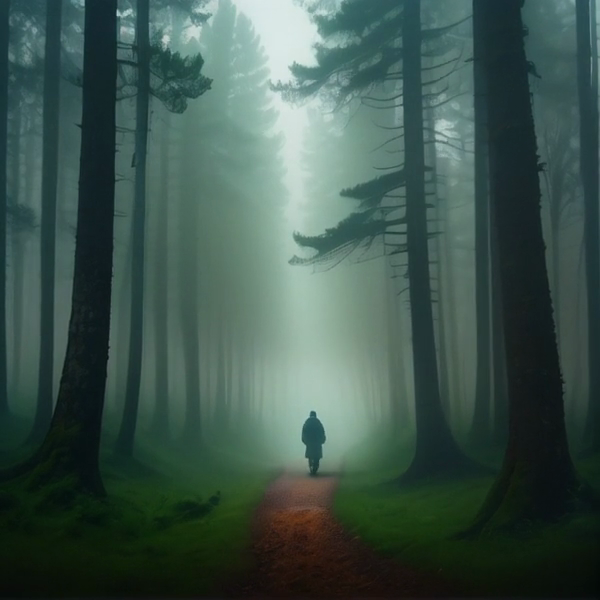}}
    \raisebox{-.5\height}{
    \begin{tikzpicture}
    \draw[->, line width=2pt] (0,0) -- (0.5,0);
    \end{tikzpicture}
    }
    \raisebox{-.5\height}{\includegraphics[width=0.25\textwidth]{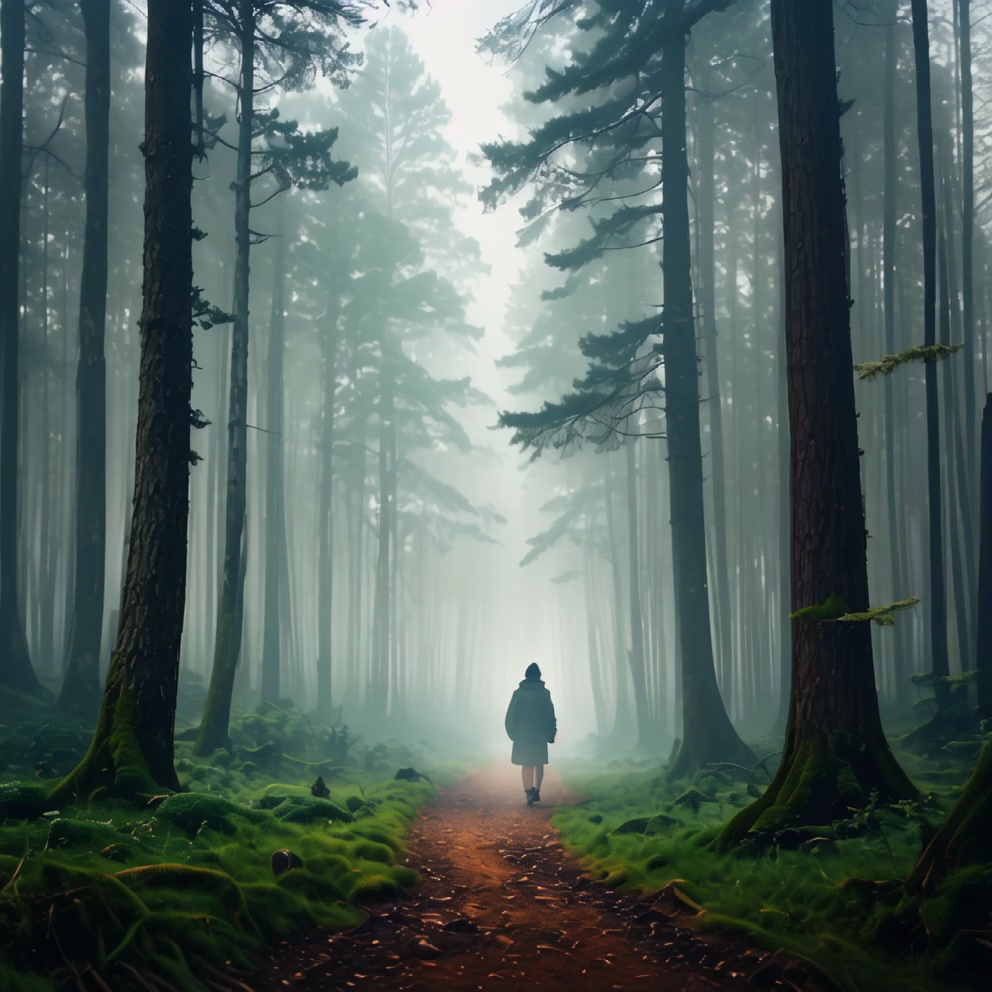}}
    \raisebox{-.5\height}{\includegraphics[width=0.25\textwidth]{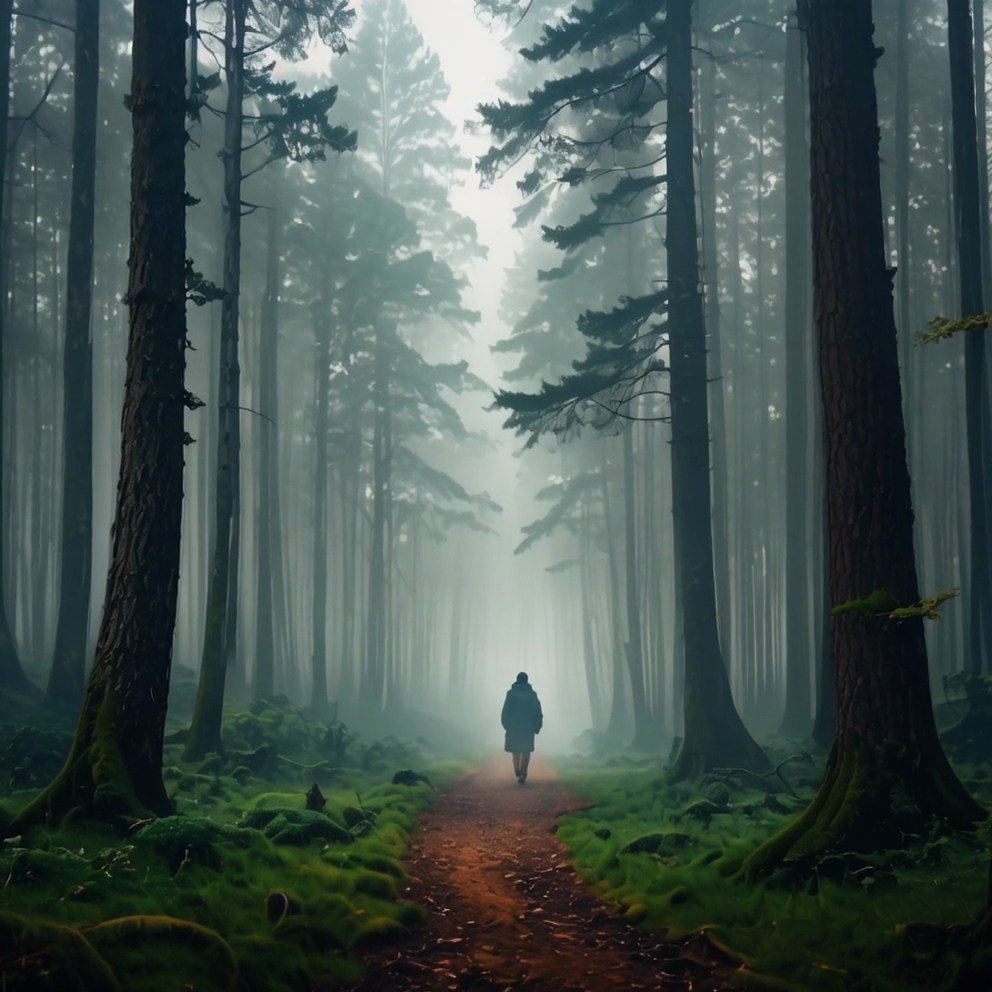}}
    \caption*{Prompt: ``A lone traveller walks in a misty forest.'' Left: low resolution video. Right: high resolution video. The tree details and scene brightness are refined by the V2V module.}
  \end{subfigure}

  \begin{subfigure}[b]{\linewidth}
    \centering
    \raisebox{-.5\height}{\includegraphics[width=0.15\textwidth]{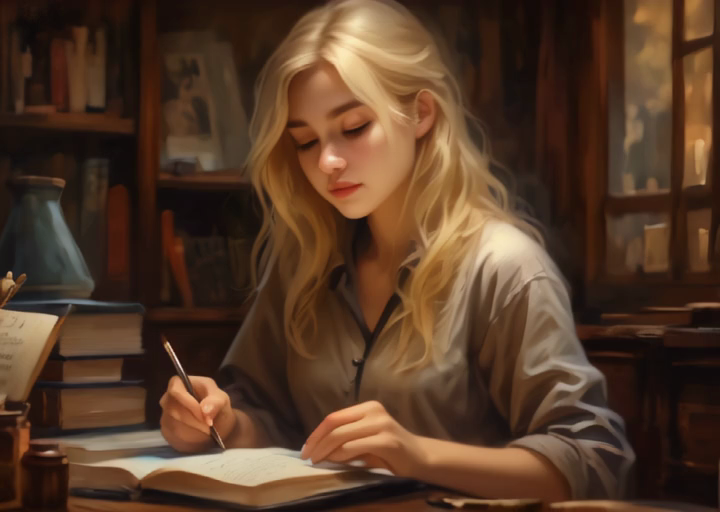}}
    \raisebox{-.5\height}{\includegraphics[width=0.15\textwidth]{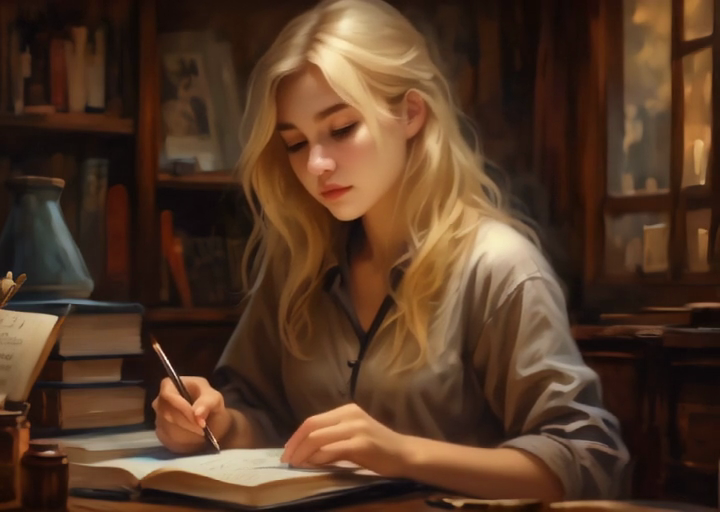}}
    \raisebox{-.5\height}{
    \begin{tikzpicture}
    \draw[->, line width=2pt] (0,0) -- (0.5,0);
    \end{tikzpicture}
    }
    \raisebox{-.5\height}{\includegraphics[width=0.25\textwidth]{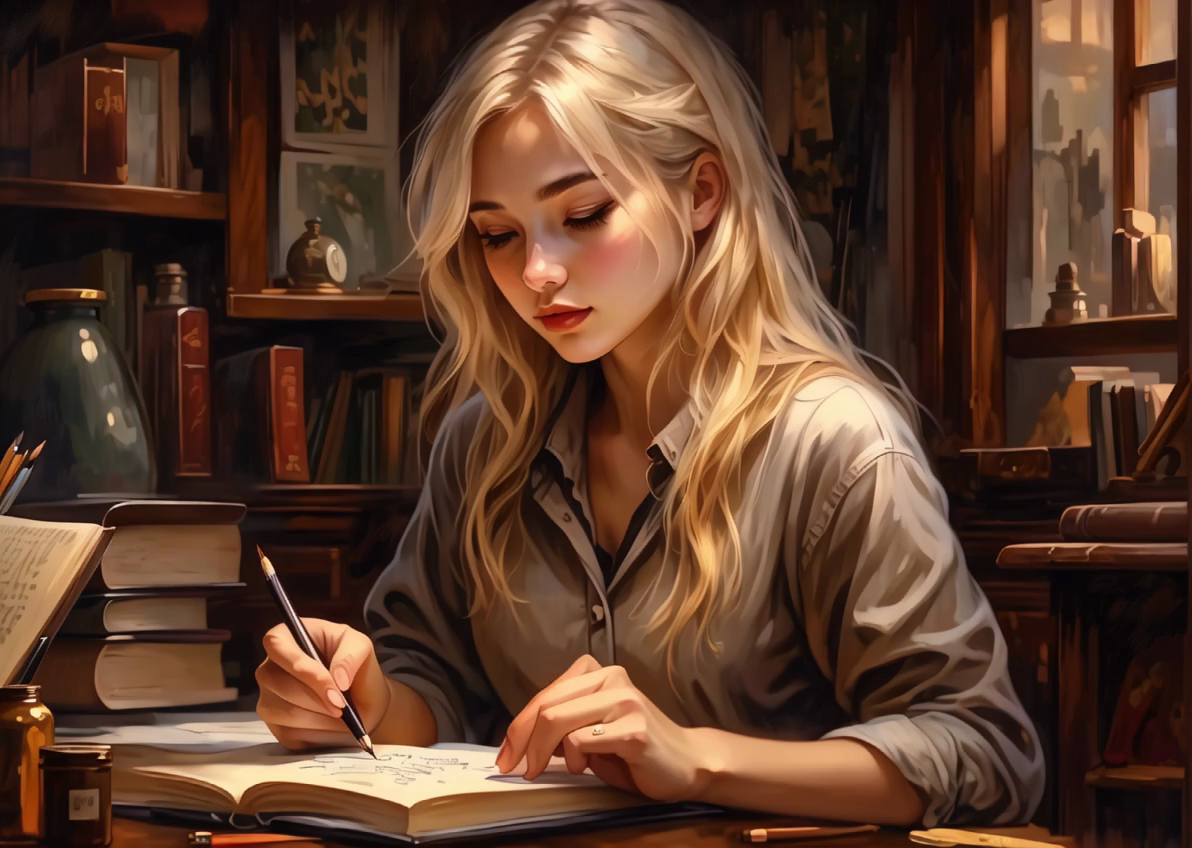}}
    \raisebox{-.5\height}{\includegraphics[width=0.25\textwidth]{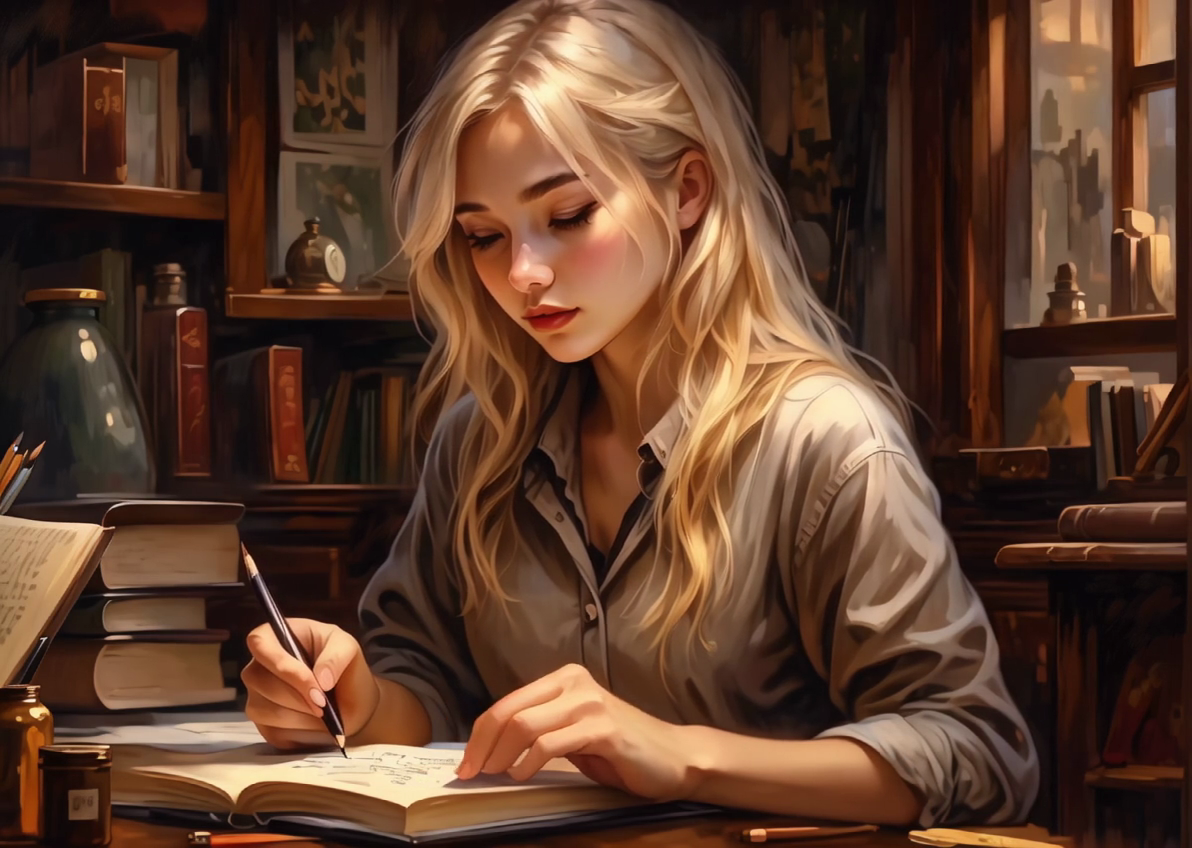}}
    \caption*{Prompt: ``A girl is writing something on a book. Oil painting style.'' Left: low resolution video. Right: high resolution video. The background and aesthetic sense are improved by the V2V module.}
  \end{subfigure}

\caption{Demonstrations of the I2V and V2V modules' capabilities to correct and refine outputs, leading to polished and visually appealing videos.}
\label{fig:refine}
\end{figure*}

Selected qualitative examples of MagicVideo-V2  are presented in Figure~\ref{fig:example}. 
For a better-viewed experience, we invite readers to watch the accompanying videos on our project website\footnote{https://magicvideov2.github.io/}. As mentioned in Section~\ref{sec:method}, the I2V and V2V modules of MagicVideo-V2 excel at rectifying and refining imperfections from the T2I module, producing smoothy and aesthetically pleasing videos. Select examples are showcased in Figure~\ref{fig:refine}.

%% file: sec/4_conclusion.tex
\section{Conclusion}
\label{sec:conclusion}

MagicVideo-V2 presents a new text-to-video generation pipeline. Our comprehensive evaluations, underscored by human judgment, affirm that MagicVideo-V2 surpasses SOTA methods.  
The modular design of MagicVideo-V2, integrating text-to-image, image-to-video, video-to-video  and video frame interpolation, provides a new strategy for generating  smooth and high-aesthetic videos.